\documentclass[final]{cvpr}
\pdfoutput=1
\usepackage{times}
\usepackage{epsfig}

\usepackage{amsmath}
\usepackage{amssymb}
\usepackage{booktabs}
\usepackage{tabulary}
\usepackage[dvipsnames,table]{xcolor}
\usepackage[british,english,american]{babel}
\usepackage{soul}
\usepackage{xspace}\xspace
\usepackage{adjustbox}
\usepackage{array}

\usepackage{dblfloatfix}
\usepackage{transparent}
\usepackage[linesnumbered,ruled,vlined]{algorithm2e}
\usepackage{listings}
\usepackage{graphicx}
\usepackage{caption}
\usepackage{subcaption}
\usepackage{amsmath,amsfonts,bm}

\def\eqref#1{equation~\ref{#1}}
\def\1{\bm{1}}

\DeclareMathAlphabet{\mathsfit}{\encodingdefault}{\sfdefault}{m}{sl}
\SetMathAlphabet{\mathsfit}{bold}{\encodingdefault}{\sfdefault}{bx}{n}

\DeclareMathOperator*{\argmin}{arg\,min}

\newcommand{\app}{\raise.17ex\hbox{$\scriptstyle\sim$}}

\newcolumntype{x}[1]{>{\centering\arraybackslash}p{#1pt}}
\newcolumntype{y}[1]{>{\raggedright\arraybackslash}p{#1pt}}
\newcolumntype{z}[1]{>{\raggedleft\arraybackslash}p{#1pt}}
\newlength\savewidth\newcommand\shline{\noalign{\global\savewidth\arrayrulewidth
  \global\arrayrulewidth 1pt}\hline\noalign{\global\arrayrulewidth\savewidth}}
\newcommand{\tablestyle}[2]{\setlength{\tabcolsep}{#1}\renewcommand{\arraystretch}{#2}\centering\footnotesize}

\SetCommentSty{mycommfont}

\SetKwInput{KwInput}{Input}                % Set the Input
\SetKwInput{KwOutput}{Output}              % set the Output

\makeatletter\renewcommand\paragraph{\@startsection{paragraph}{4}{\z@}
  {.5em \@plus1ex \@minus.2ex}{-.5em}{\normalfont\normalsize\bfseries}}\makeatother

\usepackage{comment}

\usepackage{textpos}  % for superposing a text box
\usepackage[pagebackref=true,breaklinks=true,colorlinks,bookmarks=false]{hyperref}

\def\cvprPaperID{7699} % *** Enter the CVPR Paper ID here
\def\confYear{CVPR 2021}

\newcommand{\s}[1]{{\color{magenta} [SE: #1]}}

\newcommand{\ayush}[1]{{\color{orange} [Ayush: #1]}}
\newcommand{\chenlin}[1]{{\color{blue} [CM: #1]}}

\begin{document}

%%%%%%%%% TITLE
\title{Geography-Aware Self-Supervised Learning}

\author{
Kumar Ayush\thanks{Equal Contribution. Contact: \{kayush, buzkent, chenlin\}@cs.stanford.edu}\\
% Dept. of Computer Science\\
Stanford University\\
% {\tt\small kayush@cs.stanford.edu}
\and
Burak Uzkent\footnotemark[1]\\
% Dept. of Computer Science\\
Stanford University\\
% {\tt\small buzkent@cs.stanford.edu}
\and
Chenlin Meng\footnotemark[1]\\
% Dept. of Computer Science\\
Stanford University\\
% {\tt\small chenlin@stanford.edu}
\and
Kumar Tanmay\\
% Dept. of Electrical Engg.\\
IIT Kharagpur\\
% {\tt\small kr.tanmay147@iitkgp.ac.in}
\and
Marshall Burke\\
% Dept. of Earth Science\\
Stanford University\\
% {\tt\small mburke@stanford.edu}
\and
David Lobell\\
% Dept. of Earth Science\\
Stanford University\\
% {\tt\small dlobell@stanford.edu}
\and
Stefano Ermon\\
% Dept. of Computer Science\\
Stanford University\\
% {\tt\small ermon@cs.stanford.edu}
}

% \author[1]{Kumar Ayush\thanks{Equal Contribution. Contact: \{kayush, buzkent, chenlin\}@cs.stanford.edu}}
% \author[1]{Burak Uzkent$^*$}
% \author[1]{Chenlin Meng$^*$}
% \author[2]{Kumar Tanmay}
% \author[1]{Marshall Burke}
% \author[1]{David Lobell}
% \author[1]{Stefano Ermon}

% \affil[1]{Stanford University}
% \affil[2]{IIT Kharagpur}

\maketitle

\begin{abstract}
%Self-supervised learning has shown great success on many computer vision tasks % where labeled data is lacking 
%--- 
Contrastive learning
%, one of the most popular self-supervised learning methods, 
methods have significantly narrowed the gap between supervised and unsupervised learning on computer vision tasks. 
%Despite the success, previous works on 
%Existing contrastive learning however focus mainly on conventional computer vision datasets, and applications to remote sensing data %satellite or geo-located images
%remain unexplored. 
In this paper, 
we explore their application 
%of self-supervised learning methods 
to 
%large scale 
geo-located datasets, e.g. remote sensing, where unlabeled data is often abundant but labeled data is scarce. We first show that 
due to their different characteristics, 
%on ``Functional  Map  of  the  World" (fMoW), a large-scale satellite image dataset, 
a non-trivial gap persists between contrastive and supervised learning on standard benchmarks. 
% We focus on the ``Functional  Map  of  the  World" dataset (fMoW), a large-scale dataset with satellite images. Additionally, 
To close the gap, we
propose novel training methods that exploit the spatio-temporal structure of remote sensing data. We leverage spatially aligned images over time to construct temporal positive pairs in contrastive learning and geo-location to design pre-text tasks. 
%for unsupervised learning.
%propose novel training methods that significantly improve the quality of the learned representations by outperforming the previous self-supervised learning methods approximately 10\% in terms of image classification accuracy.
%
%Finally, we utilize geo-location of the images to design an auxilary task to boost contrastive learning.
%
% Our experiments show that 
% for the previously proposed self-supervised learning methods as their supervised counterparts on many downstream tasks on ImageNet, the gap is large on remote sensing datasets. However, 
% our proposed methods is able to outperform the previous self-supervised learning methods approximately 10\% on classification tasks on fMoW. 
Our experiments show that our proposed method closes the gap between contrastive and supervised learning on image classification, object detection and semantic segmentation for remote sensing. 
Moreover, we demonstrate that the proposed method can also be applied to geo-tagged ImageNet images, improving downstream performance on various tasks.
% To further show the benefits of our method, we perform experiments on geo-tagged ImageNet images, where such method is able to significantly improve the performance of the same model on image classification \chenlin{other tasks?} by only adding an extra geo-location classification loss.
Project Webpage can be found at this link \href{https://geography-aware-ssl.github.io/}{geography-aware-ssl.github.io}.
    
\end{abstract}

\section{Introduction}
\begin{figure*}[!ht]
\centering
\begin{subfigure}[b]{0.49\textwidth}
 \centering
 \includegraphics[width=\textwidth]{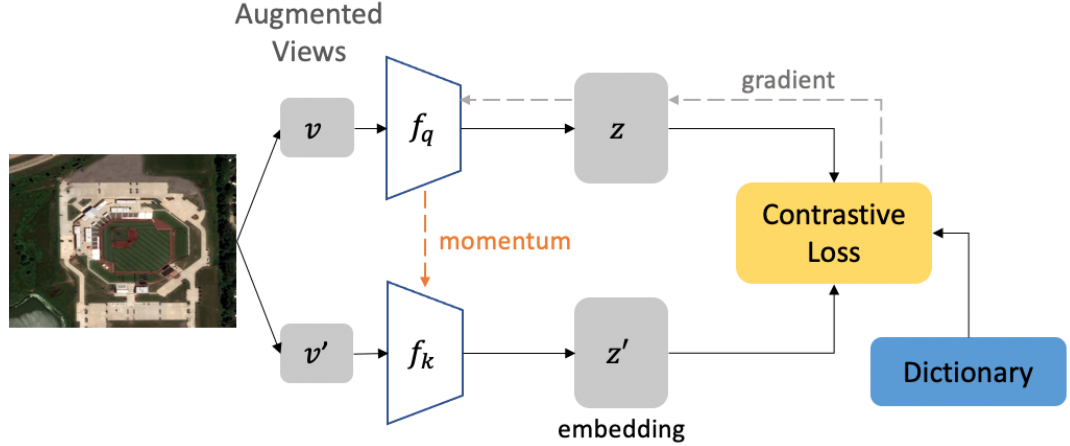}
 \label{fig:moco_base}
\end{subfigure}
~
\begin{subfigure}[b]{0.4\textwidth}
 \centering
 \includegraphics[width=\textwidth]{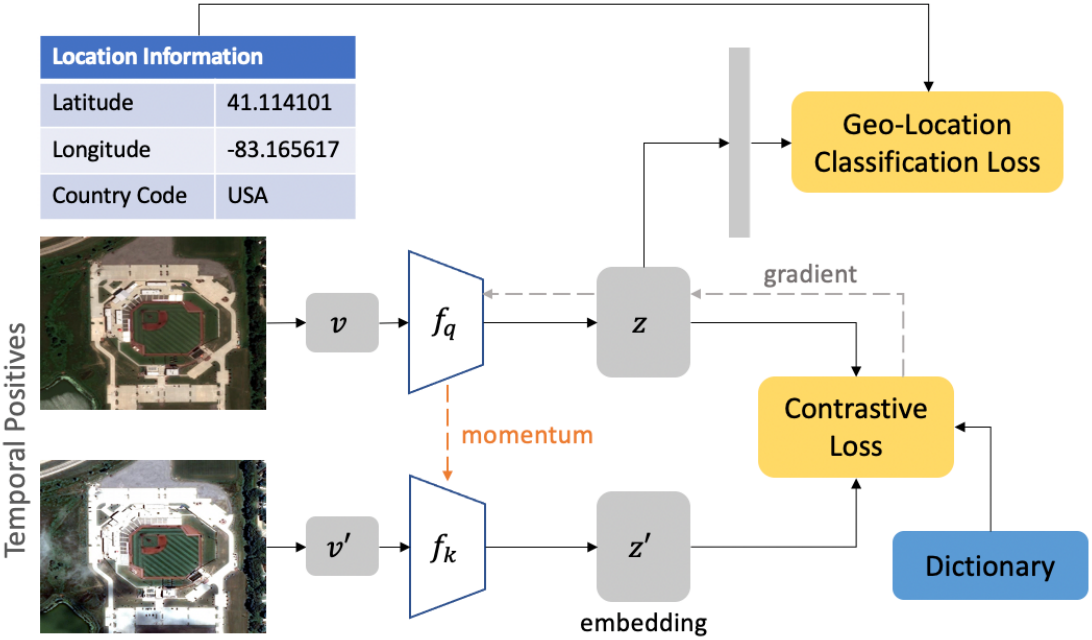}
 \label{fig:moco_ours}
\end{subfigure}
\caption{\textbf{Left} shows the original MoCo-v2~\cite{chen2020improved} framework. \textbf{Right} shows the schematic overview of our approach.}
\label{fig:approach}
\end{figure*}

Inspired by the success of self-supervised learning methods~\cite{chen2020improved,he2020momentum}, we explore their application to large-scale remote sensing datasets (satellite images) and geo-tagged natural image datasets. It has been recently shown that self-supervised learning methods perform comparably well or even better than their supervised learning counterpart on image classification, object detection, and semantic segmentation on traditional computer vision datasets~\cite{mahajan2018exploring,grill2020bootstrap,he2020momentum,chen2020improved,chen2020simple}. However, their application to remote sensing images is largely unexplored, despite the fact that collecting and labeling remote sensing images is particularly costly as annotations often require domain expertise~\cite{uzkent2019learning2,uzkent2019learning3,uzkent2019learning,lam2018xview,christie2018functional}.

In this direction, we first experimentally evaluate the performance of an existing self-supervised contrastive learning method, MoCo-v2~\cite{he2020momentum}, on remote sensing datasets, finding a performance gap with 
%between MoCo-v2~\cite{chen2020improved} and 
supervised learning using labels.
%on standard remote sensing benchmarks.
For instance, on the Functional Map of the World (fMoW) image classification benchmark~\cite{christie2018functional}, we observe an 8\% gap in top 1 accuracy between supervised and self-supervised methods.

To bridge 
this gap, 
%the gap between contrastive and supervised learning, 
we
% explore existing data augmentation methods for contrastive learning, and further 
propose geography-aware contrastive learning
%\chenlin{is "data augmentation methods" precise? that does not sound like self-supervised learning}
to leverage the spatio-temporal structure of remote sensing data. In contrast to typical computer vision images, remote sensing data are often geo-located
and might provide multiple images of the same location over time. 
%along with availability of temporal information (multiple images of the same location over time) that can be exploited.
% \s{unlike typical cv images, remote sensing data are almost always geo-located. there is also temporal structure that can be exploited, as we often get multiple images of the same location over time.}\ayush{Done}
%\s{unlike traditional .. where positive come from xxx, we }\ayush{Done.}
Contrastive methods encourage closeness of  representations of images that are likely to be semantically similar (positive pairs).
Unlike contrastive learning for traditional computer vision images where different views (augmentations) of the same image serve as a positive pair, we propose to use 
%modifying
%the contrastive learning objective to allocate 
\emph{temporal positive pairs} from spatially aligned images over time. 
Utilizing such information allows the representations to be invariant to subtle variations over time (e.g., due to seasonality). This can in turn result in more discriminative features for tasks focusing on spatial variation, such as object detection or semantic segmentation (but not necessarily for tasks involving temporal variation such as change detection). 
%useful for differentiating visually similar classes in the dataset.
%\s{why does this make sense intuitively? give intuition} \ayush{Done.}
In addition, we design a novel unsupervised learning method that leverages geo-location information, i.e., knowledge about where the images were taken. 
%\s{intuitively, just like patches from the same image are more likely to be semantically similar, images from locations that are close..}
More specifically, we consider the pretext task of predicting where in the world an image comes from,  similar to~\cite{hays2008im2gps,hays2015large}.
This can complement the information-theoretic objectives typically used by self-supervised learning methods by encouraging representations that reflect geographical information, which is often useful in remote sensing tasks~\cite{sheehan2019predicting}.  
Finally, we integrate the two proposed methods into a single geography-aware  contrastive learning objective.
%to shrink the gap with the supervised learning.

Our experiments on the functional Map of the World~\cite{christie2018functional} dataset consisting of high spatial resolution satellite images show that we improve MoCo-v2 baseline significantly. In particular, we can improve the accuracy on target applications utilizing image recognition~\cite{christie2018functional}, object detection~\cite{uzkent2020efficient,ayush2020generating}, and semantic segmentation~\cite{yao2016semantic}.
In particular, we improve it by $\sim8\%$ classification accuracy when testing the learned representations
on image classification, $\sim2\%$ AP on object detection, $\sim1\%$ mIoU on semantic segmentation, and $\sim3\%$ top-1 accuracy on land cover classification~\. Interestingly, our geography-aware learning can even outperform the supervised learning counterpart on temporal data classification by $\sim2\%$. To further demonstrate the effectiveness of our geography-aware learning approach, we extract the geo-location information of ImageNet images using FLICKR API similar to~\cite{de2019does}, which provides us with a subset of 543,435 geo-tagged ImageNet images. We extend the proposed approaches to geo-located ImageNet, and show that geography-aware learning can improve the performance of MoCo-v2 by $\sim2\%$ on image classification, showing the potential application of our approach to any geo-tagged dataset. Figure~\ref{fig:approach} shows our contributions in detail.

\section{Related Work}

% \begin{figure}[!h]
% \centering
% \includegraphics[width=\linewidth]{figures/approach.png}
% \caption{Schematic overview of our method.\burak{Notations are not consistent with the paper.}}
% \label{fig:approach}
% \end{figure}

Self-supervised methods use unlabeled data to learn representations 
%without expensive or laborious annotations such that the representations learned 
% that perform well
that are transferable to
downstream tasks (\eg image classification). Two commonly seen self-supervised methods are \emph{pre-text task} and \emph{contrastive learning}.

\paragraph{Pre-text tasks} 
Pre-text task based learning~\cite{misra2020self,vincent2008extracting,pathak2016context,zhang2017split,wang2015unsupervised,pathak2017learning} can be used to learn feature representations when data labels are not available.~\cite{gidaris2018unsupervised} rotates an image and then trains a model to predict the rotation angle.~\cite{zhang2016colorful} trains a network to perform colorization of a grayscale image. \cite{noroozi2016unsupervised} represents an image as a grid, permuting the grid and then predicting the permutation index. 
%For all these approaches, the model's activations after training can be used as  representations of the input. 
% ~\cite{kolesnikov2019revisiting} showed that performance of these approaches can be increased by using deeper and wider networks. 
In this study, we use \emph{geo-location classification} as a pre-text task, in which a deep network is trained to predict a coarse geo-location of where in the world the image might come from.
% is captured for unsupervised learning. 
%Additionally, we integrate our pre-text task into the contrastive learning framework to improve the results.
% which can then be combined with contrastive learning.
%Nevertheless, modern contrastive learning methods typically outperform the pre-text task approach. For this reason, in this study we focus on contrastive learning. \burak{should we present geo-location prediction as a pre-text task?}
%\s{yes good idea}
%\chenlin{if we mention contrastive learning in this paragraph, it might be better to move the contrastive learning paragraph before the pre-text tasks}

\paragraph{Contrastive Learning}
Recent self-supervised contrastive learning approaches such as MoCo~\cite{he2020momentum}, MoCo-v2~\cite{chen2020improved}, SimCLR~\cite{chen2020simple}, PIRL~\cite{misra2020self}, and FixMatch~\cite{sohn2020fixmatch} have demonstrated superior performance and have emerged as the fore-runner on various downstream tasks. The intuition behind these methods are to learn representations by pulling positive image pairs from the same instance closer in latent space while pushing negative pairs from difference instances further away. These methods, on the other hand, differ in the type of contrastive loss, generation of positive and negative pairs, and sampling method. 
\begin{figure*}[!h]
\centering
\includegraphics[width=0.99\textwidth]{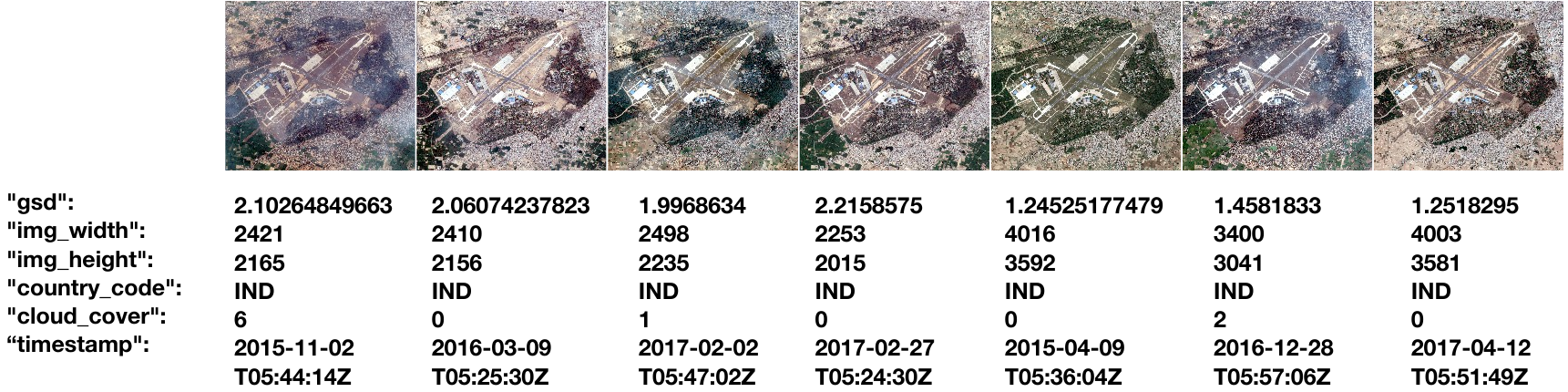}
\caption{Images over time concept in the fMoW dataset. The metadata associated with each image is shown underneath. We can see changes in contrast, brightness, cloud cover etc. in the images. These changes render spatially aligned images over time useful for constructing additional positives.}
\label{fig:images_over_time}
\end{figure*}
\begin{figure*}[!h]
\centering
\includegraphics[width=0.98\textwidth]{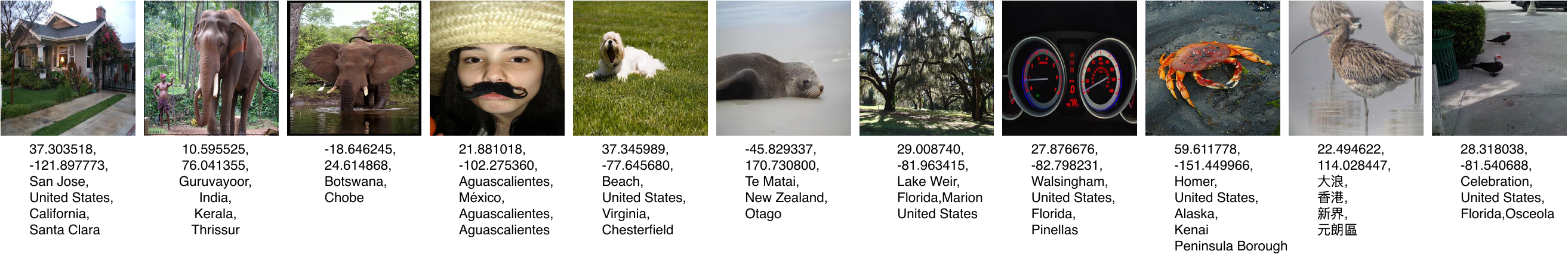}
\caption{Some examples from GeoImageNet dataset. Below each image, we list their latitudes, longitudes, city, country name. In our study, we use the latitude and longitude information for unsupervised learning. We recommend readers to zoom-in to visualize the details of the pictures.}
\label{fig:GeoImageNet_examples}
\end{figure*}
\begin{figure}[!h]
\centering
\includegraphics[width=0.23\textwidth]{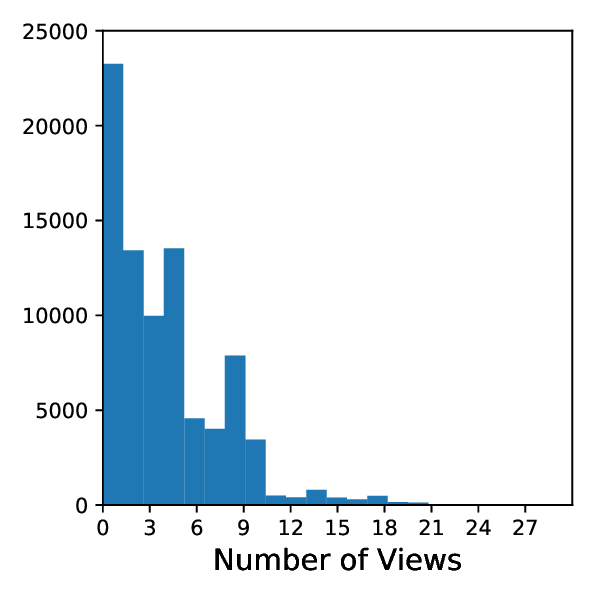}
\includegraphics[width=0.23\textwidth]{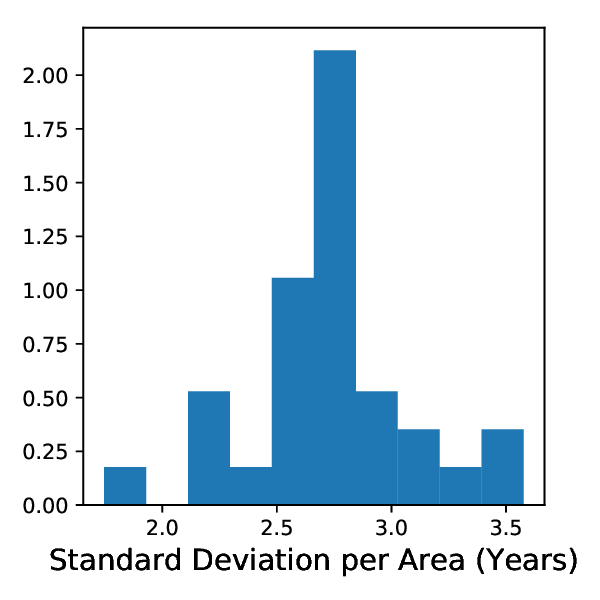}
\caption{\textbf{Left} The histogram of number of views. \textbf{Right} the histogram of standard deviation in years per area in fMoW dataset.}
\label{fig:fmow_hist_views}
\end{figure}

% The model trained on unlabeled data with contrastive loss can subsequently be fine-tuned on labeled data for particular downstream tasks.
% However, none of these methods have been applied to remote sensing data. Due to the huge domain gap between remote sensing data and natural images, existing contrastive learning approaches for natural images might not be optimal for remote sensing data. In fact, we observe a non-trivial gap between the direct application of MoCo~\cite{he2020momentum} and its supervised counterparts on remote sensing image classification, indicating that extra effort is required for adapting existing self-supervised algorithms on natural images to remote sensing domain. 
% In our experiments, we find that the vanilla MoCo-v2~\cite{he2020momentum} does not work the best for remote sensing data leaving a non-trivial gap with supervised learning counterparts to fill~\chenlin{"we find that the vanilla MoCo-v2~\cite{he2020momentum} does not work the best" shall we talk about this in related work?}~\burak{Any refence? I am not sure if this statement is true or can be justified} \ayush{Wrote in context of our work. We found that vanill Moco does not perform well} ~\burak{It actually performs well just like ImageNet, right? I think we should say that our experiments show that recent contrastive learning methods perform well on remote sensing images. To boost their performance, we provide ....}. 

Although growing rapidly in self-supervised learning area, contrastive learning methods have not been explored on large-scale remote sensing dataset. In this work, we provide a principled and effective approach for improving representation learning using MoCo-v2~\cite{he2020momentum} for remote sensing data as well geo-located conventional datasets.

\paragraph{Unsupervised Learning in Remote Sensing Images}
Unlike in traditional computer vision areas, unsupervised learning on remote sensing domain has not been studied comprehensively. Most of the studies utilize small-scale datasets specific to a small geographical region~\cite{cheriyadat2013unsupervised,li2016unsupervised,romero2015unsupervised,jean2019tile2vec,lu2017remote}, a few classes~\cite{uzkent2018tracking,mundhenk2016large} or a highly-specific modality, i.e. hyperspectral images~\cite{mou2017unsupervised,zhang2019hyperspectral}. Most of these studies focus on the UCM-21 dataset~\cite{yang2010bag} consisting of less than 1,000 images from 21 classes. A more recent study~\cite{uzkent2019learning} proposes large-scale weakly supervised learning using a multi-modal dataset consisting of satellite images and paired geo-located wikipedia articles. While being effective, this method requires each satellite image to be paired to its corresponding article, limiting the number of images that can be used.
% \burak{There is some work on unsupervised learning on remote sensing data, mostly on small scale datasets. I pretty much added all the papers. Most of them are on small datasets and region-specific, and some of them are on hyperspectral data.  \cite{romero2015unsupervised,lu2017remote,zhang2019hyperspectral,mou2017unsupervised,li2016unsupervised,jean2019tile2vec,uzkent2019learning,cheriyadat2013unsupervised,yao2016semantic,gomez2020location,lin2010joint}}

\paragraph{Geography-aware Computer Vision}
Geo-location data has been studied extensively in prior works. Most of these studies utilizes geo-location of an image as a prior to improve image recognition accuracy~\cite{tang2015improving,iso2017density,muller2018geolocation,mac2019presence,chu2019geo}. Other studies~\cite{weyand2016planet,hays2008im2gps,hays2015large,vo2017revisiting} use geo-tagged training datasets to learn how to predict the geo-location of previously unseen images at test time. In our study, we leverage geo-tag information to improve unsupervised and self-supervised learning methods.

% \burak{We can list the following studies. \cite{tang2015improving,weyand2016planet,iso2017density,muller2018geolocation,hays2008im2gps,hays2015large,vo2017revisiting,mac2019presence,chu2019geo}. Our difference is we use geo-location metadata for unsupervised learning.}
% %\section{Data Augmentation}
\section{Problem Definition}
\label{sect:problem_definition}
% \chenlin{let's use $\log, \exp$ instead of $log, exp$?}
% \chenlin{there are two same moco citations, and two same moco2 citations. check the reference}

We consider a geo-tagged visual dataset 
$\{((x_i^1, \cdots, x_i^{T_i}),\textrm{lat}_i,\textrm{lon}_i) \}_{i=1}^N$,
% where each datapoint consists of a sequence of images  $(x_i^1, \cdots, x_i^{T_i})$ of a scene over time and $\textrm{lat}_i,\textrm{lon}_i$ are the shared latitude and longitude of the $i$th image sequence.
where the $i$th datapoint consists of a sequence of images 
$\mathcal{X}_{i} = (x_i^1, \cdots, x_i^{T_i})$ at a shared location, with latitude and longitude equal to $\textrm{lat}_i,\textrm{lon}_i$ respectively, over time $t_{i}=1,...,T_i$.
%\burak{We should say that these images shares the same bounding box not just centroid.}
% , \chenlin{maybe remove $c_i$ here since it is not used in this section}as well as the  location $c_{i}$=$(\textrm{lat}_i,\textrm{lon}_i)$ where the images were taken. 
%\burak{Multi-spectral would not confuse people?}
% In the context of remote sensing, $\mathcal{X}_{i} = (x_i^1, \cdots, x_i^{T_i})$ is a sequence of images of location $(\textrm{lat}_i,\textrm{lon}_i)$ taken over a period of time. \chenlin{repeated?} When $T_i=1$ \chenlin{for all $i$?}the dataset $\{(x_i^1,\textrm{lat}_i,\textrm{lon}_i) \}_{i=1}^N$ could represent a set of individual geo-tagged images $\mathcal{X}_{i}=x_{i}^{1}$ without temporal structure.\chenlin{this is confusing because we have not defined what temporal structure is, but we are using temporal structure}\chenlin{why not define temporal structure first?}
When $T_i>1$, we refer to the dataset to have temporal information or structure. Although temporal information is often not available in natural image datasets (\eg ImageNet), it is common in remote sensing. While the temporal structure is similar to that of conventional videos, there are some key differences that we exploit in this work. First, we consider relatively short temporal sequences, where the time difference between two consecutive ``frames'' could range from months to years. 
Additionally unlike conventional videos we consider datasets where there is no viewpoint change across the image sequence. 
Given our setup, we want to obtain visual representations $z_{i}^{t_{i}}$ of images $x_{i}^{t_i}$ such that the learned representation can be transferred to various downstream tasks. We do not assume access to any labels or human supervision beyond the $\textrm{lat}_i,\textrm{lon}_i$ geo-tags. 
The quality of the representations is measured by their performance on various downstream tasks. 
%We can then perform transfer learning on different downstream tasks on remote sensing images to quantify the quality of the learned representations with our proposed approach. 
% Our primary goal is to improve existing self-supervised learning methods 
Our primary goal is to improve the performance of self-supervised learning 
%and unsupervised learning 
by utilizing the geo-coordinates and the unique temporal structure of remote sensing data.
% by utilizing the fact that we have spatially aligned images over time 
% $(x_{i}^1, \cdots, x_i^{T_i})$
%$\{x_{i}\}_{i=0}^{T_{i}}$ 
% and geo-coordinates $(\textrm{lat}_i,\textrm{lon}_i)$.
%for each unique area where we have multiple images over time $(\mathcal{X}_{i}, c_{i})$. 
%Finally, we introduce a new dataset, ImageNet with geo-location data (GeoImageNet) to improve the quality of representations by leveraging geography information. We want to note that for GeoImageNet, we do not have spatially aligned images over time, $T_{i}=0$. For this reason, we represent GeoImageNet as $(\mathcal{X},\mathcal{C})$ where $x_{i} \in \mathcal{X}$ and $c_{i} \in \mathcal{C}$. In GeoImageNet, we learn mapping $f_{q}:x_{i} \mapsto z_{i} \in \mathbb{R}^{d}$.

% \vspace{-1.0em}
\subsection{Functional Map of the World}
% \begin{figure*}[!h]
% \centering
% \includegraphics[width=0.99\textwidth]{./figures/time_series_data.pdf}
% \caption{Images over time concept in the fMoW dataset. The metadata associated with each image is shown underneath. We can see changes in contrast, brightness, cloud cover etc. in the images. These changes render spatially aligned images over time useful for constructing additional positives.}
% \label{fig:images_over_time}
% \end{figure*}
% \begin{figure*}[!h]
% \centering
% \includegraphics[width=0.98\textwidth]{./figures/geocoords_GeoImageNet.pdf}
% \caption{Some examples from GeoImageNet dataset. Below each image, we list their latitudes, longitudes, city, country name. In our study, we use the latitude and longitude information for unsupervised learning. We recommend readers to zoom-in to visualize the details of the pictures.}
% \label{fig:GeoImageNet_examples}
% \end{figure*}
% \begin{figure}[!h]
% \centering
% \includegraphics[width=0.23\textwidth]{./figures/fmow_views_hist.eps}
% \includegraphics[width=0.23\textwidth]{./figures/std_years_fmow.eps}
% \caption{\textbf{Left} The histogram of number of views. \textbf{Right} the histogram of standard deviation in years per area in fMoW dataset.}
% \label{fig:fmow_hist_views}
% \end{figure}
% \paragpraph{Functional Map of the World}
% Remote sensing area has recently seen significant advancements in publicly available large scale image datasets. Prior to 2018, there was only small-scale region-specific datasets \burak{Cite} with a few class categories. 
Functional Map of the World (fMoW) is a large-scale publicly available remote sensing dataset~\cite{christie2018functional} consisting of approximately 363,571 training images and 53,041 test images 
%from 83,000 and 15,000 unique bounding boxes 
across 62 highly granular class categories.
It provides images (temporal views) from the same location over time $(x_i^1, \cdots, x_i^{T_i})$ 
%as formulated in sect.~\ref{sect:problem_definition} 
as well as geo-location metadata $(\textrm{lat}_i,\textrm{lon}_i)$ for each image. Fig.~\ref{fig:fmow_hist_views} shows the histogram of the number of temporal views
% \chenlin{not clear what "number of views" is in this sentence}\ayush{used temporal views. does it look good now?} 
in fMoW dataset. We can see that most of the areas have multiple temporal views where $T_{i}$ can range from 1 to 21, and on average there is about 2.5-3 years of difference between the images from an area. Also, we show examples of spatially aligned images in Fig.~\ref{fig:images_over_time}.
As seen in Fig.~\ref{fig:geocoords}, fMoW is a global dataset consisting of images from seven continents which can be ideal for learning global remote sensing representations.
% for various downstream tasks where data may be restricted to particular regions. 
% \ayush{Modified the last line a bit. Does it look good and convincing.}

%Additionally, we show the examples of spatially aligned images in fig.~\ref{fig:images_over_time}.
%In our study, we utilize the \emph{temporal information} and \emph{geo-location information} of these remote sensing images to improve the quality of the learned representations and assess their performance on several downstream tasks with remote sensing images where labeling large-scale datasets are costly.

\subsection{GeoImageNet}
% \paragpraph{GeoImageNet}
\begin{comment}
\begin{figure}[!b]
\centering
\includegraphics[width=0.23\textwidth]{./figures/animals_geoimagenet.eps}
\includegraphics[width=0.23\textwidth]{./figures/countries_geoimagenet.eps}
\caption{Examples of GeoImageNet classes specific to a region in the world. \textbf{Left} shows some animals mostly found in the same region of the world and \textbf{Right} shows some classes specific to certain countries. A small portion of the Koala and African Elephant pictures have been captured in zoos in North America. We note that we do not project coordinates to the world map in this figure.}
\label{fig:cluster_examples_geoimagenet}
\end{figure}
\end{comment}
%Previously, ~\cite{de2019does} extracted geo-coordinates of ImageNet images to demonstrate the bias towards high income regions in the image recognition networks.
Following~\cite{de2019does}, we extract geo-coordinates for a subset of images in ImageNet~\cite{deng2009imagenet} using the FLICKR API.  
%to learn geography-aware representations with unsupervised learning.
%
% Another contribution of our study is adopting our methods leveraging the unique aspects of fMoW to improve learning on the ImageNet dataset. 
% The ImageNet dataset originally contains 1.2 million, 150k, and 50k images for the training, validation and test images correspondingly. 
%In this direction, we find that FLICKR API can be used to obtain the geo-location information of a subset of ImageNet images. 
More specifically, we searched for geo-tagged images in ImageNet using the FLICKR API, and were able to find 543,435 images with their associated coordinates $(\textrm{lat}_i,\textrm{lon}_i)$ across 5150 class categories. This dataset is more challenging than ImageNet-1k as it is highly imbalanced and contains about 5$\times$ more classes.  In the rest of the paper, we refer to this geo-tagged subset of ImageNet as \emph{GeoImageNet}.
% Upon publication, we will release the GeoImageNet dataset publicly for the research community.

\begin{figure}[!t]
\centering
\includegraphics[width=0.5\textwidth]{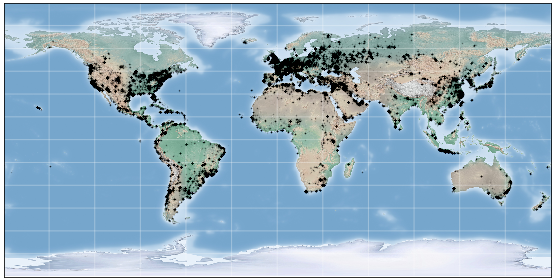}
\includegraphics[width=0.5\textwidth]{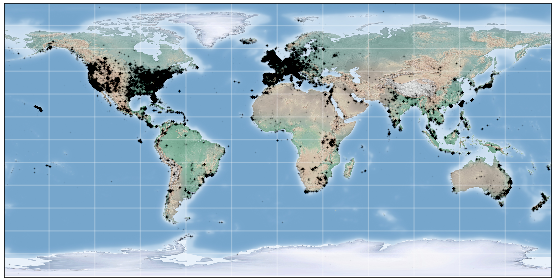}
\caption{\textbf{Top} shows the distribution of the fMoW and \textbf{Bottom} shows the distribution of GeoImageNet.}
\label{fig:geocoords}
\end{figure}

We show some examples from GeoImageNet in Fig.~\ref{fig:GeoImageNet_examples}. As shown in the figure, for some images we have geo-coordinates that can be predicted from visual cues. For example, we see that a picture of a  person with a Sombrero hat was captured in Mexico. Similarly, an Indian Elephant picture was captured in India, where there is a large population of Indian Elephants. Next to it, we show the picture of an African Elephant (which is larger in size). 
If a model is trained to predict where in the world the image was taken, it should be able to identify visual cues that are transferable to other tasks (e.g., visual cues to differentiate Indian Elephants from the African counterparts). Figure~\ref{fig:geocoords} shows the distribution of images in the GeoImageNet dataset.
%Also, a dashboard of a vehicle with indicators in miles was captured in USA which is one of the few countries officially using miles rather than kilometers. 
%In Fig.~\ref{fig:cluster_examples_geoimagenet} we show that certain animals in GeoImageNet are populated in specific regions of the world whereas some other classes are specific to certain countries.

% Using the geo-tagged ImageNet images, we are able to adopt our proposed method to Geo-tagged ImageNet to improve the MoCo-V2 method. 

% Unlike fMoW dataset, in the case of ImageNet, there exists only a single image from an area $T_i=0$. For this reason, we could not leverage the temporal positive pairs for Geo-tagged ImageNet. However, our experiments indicate that all the other techniques developed for fMoW improves the baseline MoCo-V2 on GeoImageNet.
% \burak{Show a figure for Geo-tagged ImageNet.}
% \burak{show a figure showing the distribution of the coordinates}

\section{Method}
In this section, we briefly review contrastive loss functions for unsupervised learning and detail our proposed approach to improve Moco-v2~\cite{chen2020improved}, a recent contrastive learning framework, on geo-located data.

\subsection{Contrastive Learning Framework}
%Contrastive learning~\cite{he2020momentum,chen2020simple,chen2020improved,tian2019contrastive,oord2018representation} is a framework that learns representations by contrasting semantically similar (positive) image pairs against dissimilar (negative) image pairs.
Contrastive~\cite{he2020momentum,chen2020simple,chen2020improved,tian2019contrastive,oord2018representation} methods attempt to learn a mapping $f_{q}: x_{i}^{t} \mapsto z_{i}^{t} \in \mathbb{R}^{d}$ from raw pixels $ x_{i}^{t}$ to semantically meaningful representations $z_{i}^{t}$ in an unsupervised way. 
% \s{the math is here because it'd be good to mention what is being optimized by the loss (fq), and what is the desired end product of the algorithm. but if too messy can remove the symbols}\chenlin{a tricky thing for moco is that there are two encoders $f_q$(positive pair) and $f_k$(negative pair, controlled by gradient), we might need to explain more details if we include them?}
The training objective encourages representations corresponding to pairs of images that are known a priori to be semantically similar (positive pairs) to be closer to each other than typical unrelated pairs (negative pairs).
With similarity measured by dot product, recent approaches in contrastive learning differ in the type of contrastive loss and generation of positive and negative pairs.
% Recent progress on contrastive learning focuses on the design of contrastive loss, generation of positive and negative pairs, and sampling method. 
In this work, we focus on the state-of-the-art contrastive learning framework MoCo-v2~\cite{chen2020improved}, an improved version of MoCo~\cite{he2020momentum}, 
and study improved methods for the construction of positive and negative pairs tailored to remote sensing applications.

The contrastive loss function used in the MoCo-v2 framework is InfoNCE~\cite{oord2018representation}, which is defined as follows for a given data sample:
% \s{clearer to say: given a positive pair and negative pairs, we compute representations, and optimize the paramters.. with loss.}

\begin{equation}
  L_z = -\log \dfrac{\exp(z\cdot \hat{z}/\lambda)}{\exp(z\cdot \hat{z}/ \lambda) + \sum_{j=1}^{N}  \exp(z\cdot k_{j}/ \lambda)},
\label{eq:moco}
\end{equation}
where $z$ and $\hat{z}$ are query and key representations obtained by passing the two augmented views of $x_i^{t}$ (denoted $v$ and $v'$ in Fig.~\ref{fig:approach}) through query and key encoders, $f_q$ and  $f_k$ parameterized by $\theta_q$ and $\theta_k$ respectively. Here $z$ and $\hat{z}$ form a positive pair. The $N$ negative samples, $\{k_j\}_{j=1}^{N}$, come from a dictionary of representations built as a queue. We refer readers to~\cite{he2020momentum} for details on this.
$\lambda \in \mathbb{R}^{+}$ is the temperature hyperparameter.

% where $z$ is the encoded representation of the given data sample, $\hat{z}$ is the encoded representation of the positive pair, $N$ is number of negative samples, $\{k_j\}_{j=1}^{N}$ are the encoded negative pairs and $\lambda \in \mathbb{R}^{+}$ is the temperature hyperparameter.
% In the case of the remote sensing, explain the input and output
The key idea here is to encourage representations of positive (semantically similar) pairs to be closer, and negative (semantically unrelated) pairs to be far apart as measured by dot product. The construction of positive and negative pairs plays a crucial role in this contrastive learning framework. MoCo and MoCo-v2 both use perturbations (also called ``data augmentation") from the same image to create a positive example and perturbations from different images to create a negative example. Commonly used perturbations include random color jittering, random horizontal flip, and random grayscale conversion.

\begin{figure}[!h]
\centering
\includegraphics[width=\linewidth]{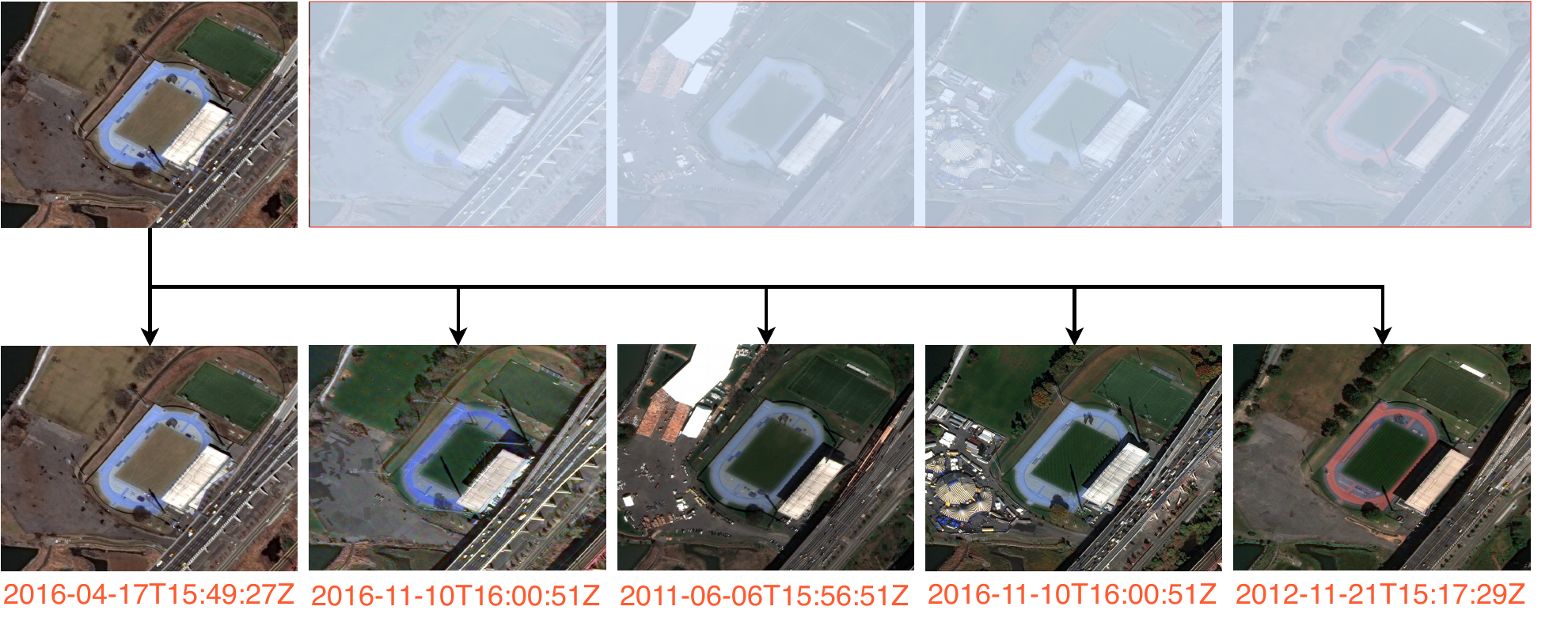}
\caption{Demonstration of temporal positives in eq.~\ref{eq:temporal_positives}. An image from an area is paired to the other images including itself from the same area captured at different time. We show the time stamps for each image underneath the images. We can see the color changes in the stadium seatings and surrounding areas.}
\label{fig:temporal_positives}
\end{figure}

\paragraph{Temporal Positive Pairs}
Different from many commonly seen natural image datasets,
% like ImageNet~\cite{deng2009imagenet}
remote sensing datasets often have extra temporal information, meaning that for a given location $(\textrm{lat}_i,\textrm{lon}_i)$, there exists
a sequence of spatially aligned images 
$\mathcal{X}_{i} = (x_i^1, \cdots, x_i^{T_i})$ over time.
% at $(\textrm{lat}_i,\textrm{lon}_i)$. 
% A very special property of the temporal information is that 
Unlike in traditional videos where nearby frames could experience large changes in content (\eg from a cat to a tree), 
%with changing camera positions, 
in remote sensing the content is often more stable across time 
due to the fixed viewpoint. 
%with camera positions pinpointed by satellites. 
For instance, a place on ocean is likely to remain as ocean for months or years, in which case satellite images taken across time at the same location should share high semantic similarities.
%since all the images are about ocean.
Even for locations where non-trivial changes do occur over time, certain semantic similarities could still remain. 
For instance, key features of a 
construction site 
are likely to remain the same even as the appearance changes due to seasonality.
%For instance, the same construction site is likely to have similar surroundings over time despite the ongoing construction.
% \chenlin{add a visualization here, maybe?}

Given these observations, it is natural to leverage temporal information for remote sensing while constructing positive or negative pairs since it can provide us with extra semantically meaningful information of a place over time. More specifically, given an image $x_i^{t_1}$ 
collected at 
%any random 
time $t_1$, we can randomly select another image $x_i^{t_2}$ that is spatially aligned with $x_i^{t_1}$ (\ie $x_i^{t_2}\in\mathcal{X}_{i}$).
%~\burak{Feels like it is conditional as given an image at a time. In practice we randomly select two indexes from a uniform distribution.}\ayush{changed a bit to address this}
We then apply perturbations (\eg random color jittering) as used in MoCo-v2 to the spatially aligned image pair $x_i^{t_1}$ and $x_i^{t_2}$, providing us with a \emph{temporal positive pair}
(denoted $v$ and $v'$ in Figure \ref{fig:approach}) 
% ($v_i^{t_1}$ and $v_i^{t_2}$)
that can be used for training the contrastive learning framework by passing them through query and key encoders, $f_q$ and $f_k$ respectively (see Fig.~\ref{fig:approach}).
%~\burak{I think we should at least say that we get views v from images using augmentations etc.}\ayush{changed to address this} 
Note that when $t_1=t_2$, the \emph{temporal positive pair} is the same as the positive pair used in MoCo-v2.
% ~\burak{you meant $t_1 = t_2$?}
%We construct negative pairs by grouping two images with different geo-locations and then applying the same perturbations used for constructing negative pairs in MoCo-v2.
%~\burak{I am not sure if this is correct. Because in practice we use the dictionary to get distances between queries and keys in the dictionary. The dictionary is updated by the representations of the sampled view from the key encoder. There is no explicit way of constructing a negative pair for an image.}\ayush{May be we can have this .... 'The negative samples come from a dictionary of encoded key representations which is built as a queue. We would refer readers to~\cite{he2020momentum} for details on this'.}

Given a data sample $x_i^{t_1}$, our TemporalInfoNCE objective function can be formulated as follows:
\begin{equation}\small
  L_{z_{i}^{t_{1}}} = -\log \dfrac{\exp(z_{i}^{t_{1}}\cdot z_{i}^{t_{2}}/\lambda)}{\exp(z_{i}^{t_{1}}\cdot z_{i}^{t_{2}}/ \lambda) + \sum\limits_{j=1}^{N} \exp(z_{i}^{t_{1}}\cdot k_{j}/ \lambda)},
  \label{eq:temporal_positives}
\end{equation}
%\burak{We should use a general term for this loss as we use this later.}
where $z_i^{t_1}$ and $z_i^{t_2}$ are the encoded representations of the randomly perturbed temporal positive pair $x_i^{t_1},x_i^{t_2}$.
%where $z_i^{t_1}$ is the encoded representation of $x_i^{t_1}$, and $z_i^{t_2}$ is the encoded representation of $x_i^{t_2}$, the temporal positive pair of $x_i^{t_1}$.
%~\burak{z is encoded representation of v?} \chenlin{we can treat data  augmentation composed with the function as encoding?}
Similar as \eqref{eq:moco}, $N$ is number of negative samples, $\{k_j\}_{j=1}^{N}$ are the encoded negative pairs and $\lambda \in \mathbb{R}^{+}$ is the temperature hyperparameter. Again, we refer readers to~\cite{he2020momentum} for details on construction of these negative pairs.
Note that compared to \eqref{eq:moco}, we use two \emph{real} images from the same area over time to create positive pairs. We believe that relying on real images for positive pairs encourages the network to learn better representations for real data than the one relying on synthetic images.
On the other hand, our objective in \eqref{eq:temporal_positives} enforces the representations to be invariant to changes over time. Depending on the target task, such inductive bias can be desirable or undesirable. For example, for a change detection task, learning representations that are highly sensitive to temporal changes can be more preferable. However, for image classification or object detection task, learning temporally invariant features should not degrade the downstream performance.
\subsection{Geo-location Classification as a Pre-text Task}
%In the previous section, we leverage the spatially aligned remote sensing images to construct temporal positive pairs in MoCo-v2 objective. 
% Geo-location metadata has been previously explored by different studies to improve image recognition accuracy~\cite{tang2015improving,mac2019presence,chu2019geo,lin2010joint,christie2018functional}, image retrieval~\cite{gomez2020location} and predict geo-location of the images missing meta-data~\cite{hays2008im2gps,hays2015large}. 
% In this study, we utilize geo-location metadata for unsupervised learning.

% In this direction, we use geo-locations of the images to design a pre-text task. 
% %\chenlin{not clear what the second method is}
% Pre-text tasks have been commonly used for unsupervised learning and some of them can be listed as solving jigsaw puzzle~\cite{noroozi2016unsupervised}, rotation angle prediction~\cite{gidaris2018unsupervised}, and colorization of a gray-scale image~\cite{zhang2016colorful}. All these methods apply a pre-defined transformation to the images and tasks the network to predict the transformation. 
In this section, we explore another aspect of remote sensing images, \emph{geolocation metadata}, to further improve the quality of the learned representations. In this direction, we design a pre-text task for unsupervised learning. In our pre-text task, we cluster the images
%, $\mathcal{X}$, 
in the dataset using their coordinates $(\textrm{lat}_{i}, \textrm{lon}_i)$. We use a clustering method to construct $K$ clusters and assign an area with coordinates $(\textrm{lat}_{i}, \textrm{lon}_i)$ a categorical geo-label $c_{i} \in \mathcal{C} = \{1, \cdots, K\}$.
%to construct $(\mathcal{X}, \mathcal{C})$. 
Using the cross entropy loss function, we then optimize a  geo-location predictor network $f_{c}$ as
%$L_{g}(f_{c}(x_{i}^{t}), c_{i})$.
%defined as:
% \begin{equation}
%     L_{g}(x_{i}^{t_i}, c_{i}) = -\sum_{i=1}^{M} p(c_{i})\log(f_{c}(x_{i}^{t_i})),
% \end{equation}
\begin{equation}
    L_{g} = \sum_{k=1}^{K}-p(c_{i}=k)\log(\hat{p}(c_{i}=k|f_{c}(x_{i}^{t})),
\end{equation}
%\chenlin{the notation is not clear, what is $f_{c}(x_{i}^{t})$?}
%where we predict the geo-labels as $f_{c}:x_{i}^{t_i} \mapsto \hat{c}_{i} \in \mathcal{R}^{1}$. 
where $p$ represent the probability of the cluster k representing the true cluster and $\hat{p}$ represents the predicted probabilities for $K$ clusters.
In our experiments, we represent $f_{c}$ with a CNN parameterized by $\theta_c$. With this objective, our goal is to learn location-aware representations that can potentially transfer well to different downstream tasks.

\subsection{Combining Geo-location and Contrastive Learning Losses}
In the previous section, we designed a pre-text task leveraging the geo-location meta-data of the images to learn location-aware representations in a standalone setting. In this section, we combine geo-location prediction and contrastive learning tasks in a single objective to improve the contrastive learning-only and geo-location learning-only tasks. In this direction, we first integrate the geo-location learning task into the contrastive learning framework using the cross-entropy loss function %$L_g(f_c(z_{i}^{t}),c_i)$
where the geo-location prediction network uses features $z_{i}^{t}$ from the query encoder as
\begin{equation}
     L_{g} = -\sum_{i=1}^{K} p(c_{i}=k)\log(\hat{p}(c_{i}=k|f_{c}(z_{i}^{t})).
\end{equation}
% \begin{equation}
%      L_{g}(z_{i}^{t_i}, c_{i}) = -p(c_{i})\log(f_{c}(z_{i}^{t_i}))
% \end{equation}
%In the geo-location prediction only task, we use a CNN to learn representations as $f_{c}:x_{i}^{t_i} \mapsto \hat{c}_{i} \in \mathcal{R}^{1}$ where $f_c$ is parameterized by a CNN. 
In the combined framework (see Fig.~\ref{fig:approach}),
%design it as  $f_{c}:z_{i}^{t_i} \mapsto \hat{c}_{i} \in \mathcal{R}^{1}$ where 
$f_c$ is represented by a linear layer parameterized by $\theta_c$.
Finally, we propose the objective for joint learning as the linear combination of TemporalInfoNCE and geo-classification loss with coefficients
% \begin{align}
%     \argmin_{\theta_{q},\theta_{k},\theta_{c}} L_{\textrm{final}}(x_{i}^{t_i},c_{i},\theta_{q},\theta_{k},\theta_{c}) = \alpha L_{t}(x_{i}^{t_i},\theta_{q},\theta_{k}) + \nonumber \\ \beta L_{g}(z_{i}^{t_i},c_{i},\theta_{c})
% \end{align}
% \begin{align}
%     \argmin_{\theta_{q},\theta_{k},\theta_{c}} L_{\textrm{final}}(z_i^{t_1},z_i^{t_2},c_{i}) = \alpha L_{t}(z_{i}^{t_1},z_{i}^{t_2}) + \nonumber \\ \beta L_{g}(f_{c}(z_i^{t_1}),c_{i})
% \end{align}
$\alpha$ and $\beta$ representing the importance of contrastive learning and geo-location learning losses as 
\begin{equation}
    \argmin_{\theta_{q},\theta_k,\theta_c}L_f = \alpha L_{z^{t_{1}}} + \beta L_g.
\end{equation}
%We show the pseudocode for the combined method in Algorithm~\ref{alg:algorithm}. 
%The objective is jointly optimized as a function of the query and key encoder parameters $\theta_q$, $\theta_k$ and $\theta_c$.
By combining two tasks, we learn representations to jointly maximize agreement between spatio-temporal positive pairs, minimize agreement between negative pairs and predict the geo-label of the images from the positive pairs.

\section{Experiments}
In this study, we perform unsupervised representation learning on fMoW and GeoImageNet datasets. We then evaluate the learned representations/pre-trained models on a variety of downstream tasks including image recognition, object detection and semantic segmentation benchmarks on remote sensing and conventional images.

\begin{figure}[!h]
\centering
\includegraphics[width=0.198\textwidth]{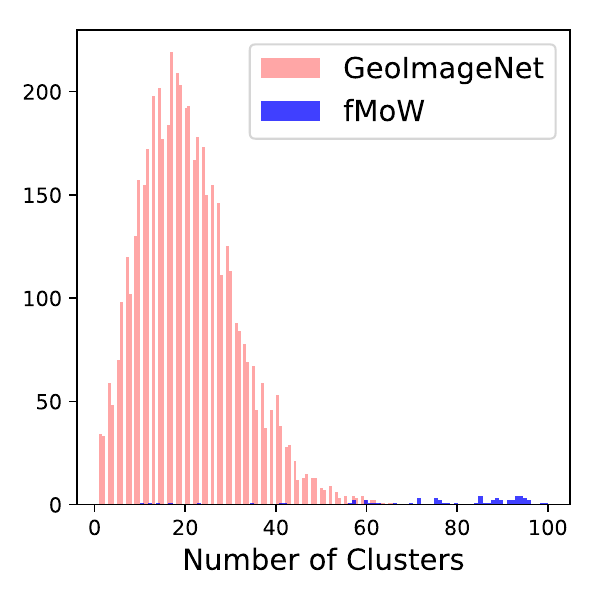}
\includegraphics[width=0.193\textwidth]{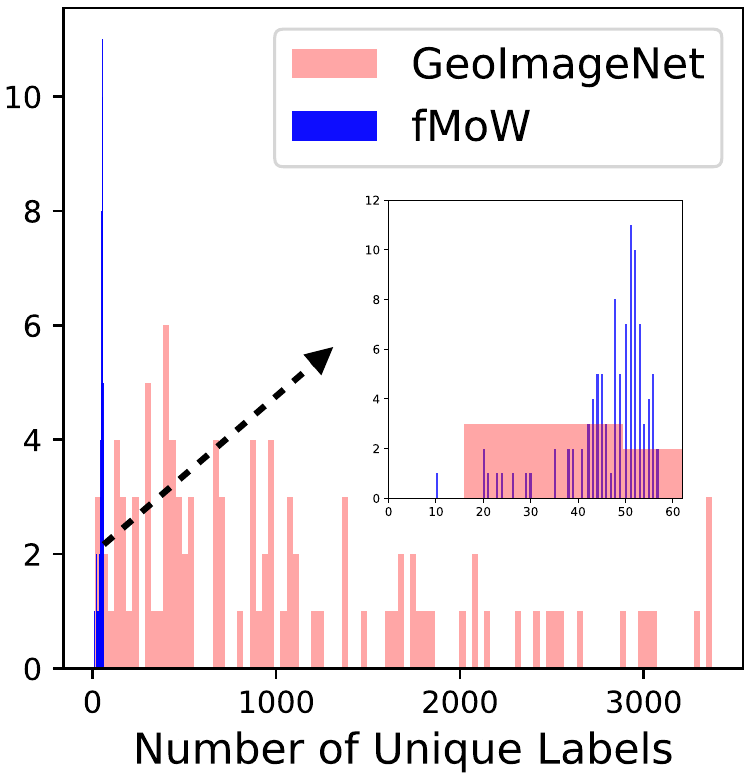}
\caption{\textbf{Left} shows the number of clusters per label and \textbf{Right} shows the number of unique labels per cluster in fMoW and GeoImageNet. Labels represent the original classes in fMoW and GeoImageNet.
% \chenlin{maybe clarify what the y labels are?}
}
\label{fig:unique_clusters_per_label}
\end{figure}
\begin{figure}[!b]
\centering
\includegraphics[width=\linewidth]{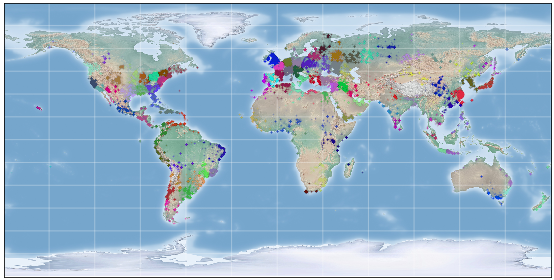}
\includegraphics[width=\linewidth]{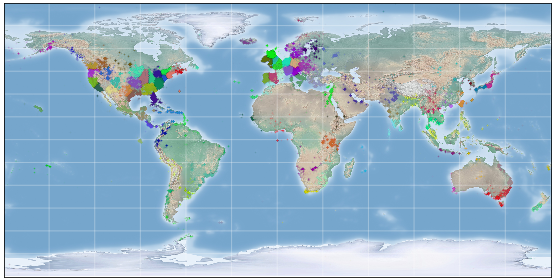}
\caption{\textbf{Top} and \textbf{Bottom} show the distributions of the fMoW and GeoImageNet clusters.}
\label{fig:clustered_dataset}
\end{figure}

\paragraph{Implementation Details for Unsupervised Learning}
\emph{For contrastive learning}, similar to MoCo-v2~\cite{chen2020improved}, we use ResNet-50 to paramaterize the query and key encoders, $f_{q}$ and $f_{k}$, in all experiments. We use following hyper-parameters in the MoCo-v2 pre-training step: learning rate of 1e-3, batch size of 256, dictionary queue of size 65536, temperature scaling of 0.2 and SGD optimizer. We use similar setups for both fMoW and GeoImageNet datasets. Finally, for each downstream experiment, we report results for the representations learned after 200 epochs. 
%We believe this is critical as MoCo-v2 performs better on downstream tasks when trained more number of epochs.

\emph{For geo-location classification task}, we run k-Means clustering algorithm to cluster fMoW and GeoImageNet into $K=100$ geo-clusters given their latitude and longitude pairs. We show the clusters in Fig.~\ref{fig:clustered_dataset}. As seen in the figure, while both datasets have similar clusters there are some differences, particularly in North America and Europe. In Fig.~\ref{fig:unique_clusters_per_label} we analyze the clusters in GeoImageNet and fMoW. 
% \s{dont understand these histograms. why should they be comparable if they are compleetly different datasets?}
The figure shows that the number of clusters per class on GeoImageNet tend to be skewed towards smaller numbers than fMoW whereas the number of unique classes per cluster on GeoImageNet has more of a uniform distribution. For fMoW, we can conclude that each cluster contain samples from most of the classes.
Finally, when adding the geo-location classification task into the contrastive learning we tune $\alpha$ and $\beta$ to be 1.

\paragraph{Methods} We compare our unsupervised learning approach to \textbf{supervised learning} for image recognition task. For object detection, and semantic segmentation we compare them to pre-trained weights obtained using (a) \textbf{supervised learning}, and (b) \textbf{random initilization} while fine-tuning on the target task dataset. Finally, for ablation analysis we provide results using different combinations of our methods. When appending only geo-location classification task or temporal positives into \textbf{MoCo-v2} we use \textbf{MoCo-v2+Geo} and \textbf{MoCo-v2+TP}. When adding both of our approaches into \textbf{MoCo-v2} we use \textbf{MoCo-v2+Geo+TP}.
% \paragraph{Metrics}
% In our study, for image recognition task, we use the top-1 and top-5 classification accuracies. Meanwhile, for object detection and semantic segmentation experiments we use the average precision ($AP_{50}$) and intersection-over-union (mIoU) metrics. \burak{Does not add anything extra}

\subsection{Experiments on fMoW}
We first perform experiments on fMoW image recognition task. Similar to the common protocol of evaluating unsupervised pre-training methods~\cite{chen2020improved,he2020momentum}, we freeze the features and train a supervised linear
classifier. However, in practice, it is more common to finetune the features end-to-end in a downstream task. For completeness and a better comparison, we report end-to-end finetuning results for the 62-class fMoW classification as well. We report both top-1 accuracy and F1-scores for this task.

\definecolor{Gray}{gray}{0.5}
\newcommand{\supimgnetpret}{\tablestyle{1pt}{1} \begin{tabular}{z{21}y{26}} \multicolumn{2}{c}{\small Sup. Learning (IN wts. init.)} \end{tabular}}
\newcommand{\supimgnet}{\tablestyle{1pt}{1} \begin{tabular}{z{21}y{26}} \multicolumn{2}{c}{\small Sup. Learning (Scratch)} \end{tabular}}
\newcommand{\geoloc}{\tablestyle{0pt}{1} \begin{tabular}{z{21}y{26}} \multicolumn{2}{c}{\small \textbf{Geoloc. Learning}} \end{tabular}}
\newcommand{\baselineinpret}{\tablestyle{0pt}{1} \begin{tabular}{z{21}y{26}} \multicolumn{2}{c}{\small \textbf{MoCo-V2 (pre. on IN)}} \end{tabular}}
\newcommand{\baseline}{\tablestyle{0pt}{1} \begin{tabular}{z{21}y{26}} \multicolumn{2}{c}{\small \textbf{MoCo-V2}} \end{tabular}}
\newcommand{\baselinegeo}{\tablestyle{0pt}{1} \begin{tabular}{z{21}y{26}} \multicolumn{2}{c}{\small \textbf{MoCo-V2+Geo}} \end{tabular}}
\newcommand{\baselinetp}{\tablestyle{0pt}{1} \begin{tabular}{z{21}y{26}} \multicolumn{2}{c}{\small \textbf{MoCo-V2+TP}} \end{tabular}}
\newcommand{\final}{\tablestyle{0pt}{1} \begin{tabular}{z{21}y{26}} \multicolumn{2}{c}{\small \textbf{MoCo-V2+Geo+TP}} \end{tabular}}

% Old Table
% \begin{table}[!h]
% \resizebox{0.98\columnwidth}{!}{%
% \begin{tabular}{@{}c|c|c|c@{}}
% \toprule
%  & \textbf{Backbone} & \textbf{\begin{tabular}[c]{@{}c@{}}Accuracy $\uparrow$\\ (100 Epochs)\end{tabular}} & \textbf{\begin{tabular}[c]{@{}c@{}}Accuracy $\uparrow$\\ (200 Epochs)\end{tabular}} \\ \midrule
% \supimgnet* & ResNet50 & 69.05 & 69.05 \\
% \hline
% \geoloc* & ResNet50 & 52.40 & 52.40 \\
% \hline
% \baseline & ResNet50 & 58.32 & 60.69 \\
% \baselinegeo & ResNet50 & 63.65 & 64.07 \\
% \baselinetp & ResNet50 & \cellcolor{blue!25}\textbf{67.15} & \cellcolor{blue!25}\textbf{68.32} \\
% \final & ResNet50 & 65.77 & 66.33 \\
% \end{tabular}}
% \caption{Experiments on fMoW on classifying single images. * indicates a model trained up to epoch with the highest accuracy on the validation set. We use the same set up for Sup. Learning and Geoloc. Learning in the remaining experiments.}
% \label{tab:fmow_single}
% \end{table}

\begin{table}[!h]
\resizebox{0.98\columnwidth}{!}{%
\begin{tabular}{@{}c|c|c|c@{}}
\toprule
 & \textbf{Backbone} & \textbf{\begin{tabular}[c]{@{}c@{}}F1-Score $\uparrow$\\ (Frozen/Finetune) \end{tabular}} & \textbf{\begin{tabular}[c]{@{}c@{}}Accuracy $\uparrow$\\ (Frozen/Finetune) \end{tabular}} \\ \midrule
\supimgnetpret* & ResNet50 & -/64.72 & -/69.07 \\
\supimgnet* & ResNet50 & -/64.71 & -/69.05 \\
\hline
\geoloc* & ResNet50 & 48.96/52.23 & 52.40/56.59 \\
\hline
\baselineinpret & ResNet50 & 31.55/57.36 & 37.05/62.90 \\
\hline
\baseline & ResNet50 & 55.47/60.61 & 60.69/64.34 \\
\baselinegeo & ResNet50 & 61.60/66.60 & 64.07/69.04 \\
\baselinetp & ResNet50 & \cellcolor{blue!25}\textbf{64.53/67.34} & \cellcolor{blue!25}\textbf{68.32/71.55} \\
\final & ResNet50 & 63.13/66.56 & 66.33/70.60 \\
\end{tabular}}
\caption{Experiments on fMoW on classifying single images. * indicates a model trained up to epoch with the highest accuracy on the validation set. We use the same set up for Sup. Learning and Geoloc. Learning in the remaining experiments. \textbf{Frozen} corresponds to linear classification on frozen features. \textbf{Finetune} corresponds to end-to-end finetuning results for the fmow classification.}
\label{tab:fmow_single}
\end{table}

\paragraph{Classifying Single Images}
%\chenlin{the title is very confusing, people would assume that the classification is performed on one image}
In Table~\ref{tab:fmow_single}, we report the results on single image classification on fMoW. We would like to highlight that in this case we classify each image individually. 
%by performing the following mapping $x_{i}^{t} \mapsto \hat{y}_{i}^{t} \in \mathbb{R}^{1}$. 
In other words, we do not use the prior information that multiple images over the same area $(x_{i}^{1}, x_{i}^{2}, \dots, x_{i}^{T_{i}})$ have the same labels $(y_{i}, y_{i}, \dots, y_{i})$. For evaluation, we use 
%compare the predictions 
%$\hat{y}_{i}^{t}$ to $y_{i}^{t}$ 
% on 
53,041 images.
%\chenlin{"classify each image independently" is confusing, people would assume that}
Our results on this task (linear classification on frozen features) show that MoCo-v2 performs reasonably well on a large-scale dataset with $60.69\%$ accuracy, $8\%$ less than the supervised learning methods. \emph{Sup. Learning (IN wts. init.)} and \emph{Sup. Learning (Scratch)} correspond to supervised learning method starting from imagenet pre-trained weights and random weights respectively. This result aligns with MoCo-v2's performance on the ImageNet dataset~\cite{chen2020improved}. Next, by incorporating geo-location classification task into MoCo-v2, we improve by $3.38\%$ in top-1 classification accuracy. We further improve the results to $68.32\%$ using temporal positives, bridging the gap between the MoCo-v2 baseline and supervised learning to less than $1\%$. However, when we perform end-to-end finetuning for the classification task, we observe that our method surpasses the supervised learning methods by more than $2\%$. For completeness, we also include results for MoCo-v2 pre-trained on Imagenet dataset (4th row in Table~\ref{tab:fmow_single}) and find that the distribution shift between Imagenet and downstream dataset leads to suboptimal performance.
% \vspace{-1em}
\paragraph{Classifying Temporal Data}
%\chenlin{maybe change the title to include "classification", reasoning is not very clear}
In the next step, we change how we perform testing across multiple images over an area at different times. In this case, we predict labels from images over an area i.e. 
make a prediction 
%perform $x_{i}^{t} \mapsto \hat{y}_{i}^{t} \in \mathbb{R}^{1}$ 
for each $t \in \{1,\dots,T_i\}$,  and average the predictions from that area. We then use the most confident class prediction to get area-specific class predictions.  %$\mathcal{\hat{Y}}_{i}$. 
In this case, we evaluate the performance on 11,231 unique areas that are represented by multiple images at different times.  
%by comparing predictions $\mathcal{\hat{Y}}_{i}$ to ground truth $\mathcal{Y}_{i}$. 
Our results in Table~\ref{tab:fmow_temporal} show that doing area-specific inference improves the classification accuracies by $4$-$8\%$ over image-specific inference. Even incorporating temporal positives, we can improve the accuracy by $6.1\%$ by switching from image classification to temporal data classification.
%It is interesting that by adding temporal positives and learning temporal invariance, the improvement shrinks from $7.95\%$ to $6.1\%$. We justify that happens because of the sequences with object-level changes. 
% Class-level accuracies for each method on image and temporal data classification tasks are shown in \textbf{Appendix}. 
%where we show that with temporal positives the accuracy goes down by $2\%$ when switching from image-specific to area-specific inference. O
Overall, our methods outperform the baseline Moco-v2 by $4$-$6\%$ and supervised learning by $1$-$2\%$. Here we only report temporal classification on top of frozen features. 

\begin{table}[!h]
\resizebox{0.98\columnwidth}{!}{%
\begin{tabular}{@{}c|c|c|c@{}}
\toprule
 & \textbf{Backbone} & \textbf{\begin{tabular}[c]{@{}c@{}}F1-Score $\uparrow$\end{tabular}} & \textbf{\begin{tabular}[c]{@{}c@{}}Accuracy $\uparrow$\end{tabular}} \\ \midrule
\smaller 
\supimgnetpret* & ResNet50 & 68.72 (+4.01) & 73.22 (+4.15) \\
\supimgnet* & ResNet50 & 68.73 (+4.02) & 73.24 (+4.19) \\
\hline
\geoloc* & ResNet50 & 52.01 (+3.05) & 56.12 (+3.72) \\
\hline
\baselineinpret & ResNet50 & 35.93 (+4.38) & 42.56 (+5.51) \\
\hline
\baseline & ResNet50 & 63.96 (+8.49) & 68.64 (+7.95) \\
\baselinegeo & ResNet50 & 66.93 (+5.33)& 70.48 (+6.41) \\
\baselinetp & ResNet50 & \cellcolor{blue!25} \textbf{70.11 (+5.58)} & \cellcolor{blue!25} \textbf{74.42 (+6.10)} \\
\final & ResNet50 & 69.56 (+6.43) & 72.76 (+6.43) \\
\end{tabular}}
\caption{Experiments on fMoW on classifying temporal data. In the table, we compare the results to the ones on single image classification. Here we present results corresponding to linear classification on frozen features only. 
% End-to-end finetuning results are present in \textbf{Appendix}.
}
\label{tab:fmow_temporal}
\end{table}

\subsection{Transfer Learning Experiments}
Previously, we performed pre-training experiments on fMoW dataset and quantified the quality of the representations by supervised training a linear layer for image recognition on fMoW. In this section, we perform transfer learning experiments on different low level tasks. 
% \vspace{-1em}
\subsubsection{Object Detection}
For object detection, we use the xView dataset~\cite{lam2018xview} consisting of high resolution satellite images captured with similar sensors to the ones in the fMoW dataset. The xView dataset consists of 846 very large ($\sim$2000$\times$2000 pixels) satellite images with bounding box annotations for 60 different class categories including airplane, passenger vehicle, maritime vessel, helicopter etc.
% \vspace{-0.2em}
\paragraph{Implementation Details}
We first divide the set of large images into 700 training and 146 test images. Then, we process the large images to create 416$\times$416 pixels images by randomly sampling the bounding box coordinates of the small image and we repeat this process 100 times for each large image. In this process, we ensure that there is less than $25\%$ overlap between any two bounding boxes from the same image. We then use RetinaNet~\cite{lin2017focal} with pre-trained ResNet-50 backbone and fine-tune the full network on the xView training set. To train RetinaNet, we use learning rate of 1e-5 and a batch size of 4 and Adam optimizer.
\vspace{-0.2em}
\paragraph{Qualitative Analysis}
Table~\ref{tab:xview} shows the object detection performance on the xView test set.  We achieve the best results with the addition of temporal positive pair, and geo-location classification pre-text task into MoCo-v2. With our final model, we can outperform the randomly initialized weights by $7\%$ AP and the supervised learning on the fMoW by $3.3\%$ AP.

\definecolor{Gray}{gray}{0.5}
\newcommand{\randinit}{\tablestyle{1pt}{1} \begin{tabular}{z{21}y{26}} \multicolumn{2}{c}{\demph{Random Init.}} \end{tabular}}
\newcommand{\iminit}{\tablestyle{1pt}{1} \begin{tabular}{z{21}y{26}} \multicolumn{2}{c}{\demph{Imagenet Init.}} \end{tabular}}
\newcommand{\mocobs}{\tablestyle{0pt}{1} \begin{tabular}{z{21}y{26}} \multicolumn{2}{c}{\textbf{MoCo-V2}} \end{tabular}}
\newcommand{\mocobsrs}{\tablestyle{0pt}{1} \begin{tabular}{z{21}y{26}} \multicolumn{2}{c}{\textbf{MoCo-V2-Geo}} \end{tabular}}
\newcommand{\mocobsrss}{\tablestyle{0pt}{1} \begin{tabular}{z{21}y{26}} \multicolumn{2}{c}{\textbf{MoCo-V2-TP}} \end{tabular}}
\newcommand{\mocobsrsst}{\tablestyle{0pt}{1} \begin{tabular}{z{21}y{26}} \multicolumn{2}{c}{\textbf{MoCo-V2-Geo+TP}} \end{tabular}}
\newcommand{\mocobsrsstg}{\tablestyle{0pt}{1} \begin{tabular}{z{21}y{26}} \multicolumn{2}{c}{\textbf{MoCo-V2-TG(448)}} \end{tabular}}
% ------------------------------------------------
\newcommand{\demph}[1]{\textcolor{Gray}{#1}}
\newcommand{\std}[1]{{\fontsize{5pt}{1em}\selectfont ~~$_\pm$$_{\text{#1}}$}}

\definecolor{Highlight}{HTML}{39b54a}  % green

\renewcommand{\hl}[1]{\textcolor{Highlight}{#1}}

\newcommand{\res}[3]{
\tablestyle{1pt}{1}
\begin{tabular}{z{16}y{18}}
{#1} &
\fontsize{7.5pt}{1em}\selectfont{~(${#2}${#3})}
\end{tabular}}

\newcommand{\reshl}[3]{
\tablestyle{1pt}{1} 
\begin{tabular}{z{16}y{18}}
{#1} &
\fontsize{7.5pt}{1em}\selectfont{~\hl{(${#2}$\textbf{#3})}}
\end{tabular}}

\newcommand{\resrand}[2]{\tablestyle{1pt}{1} \begin{tabular}{z{16}y{18}} \demph{#1} & {} \end{tabular}}
\newcommand{\ressup}[2]{\tablestyle{1pt}{1} \begin{tabular}{z{16}y{18}} {#1} & {} \end{tabular}}

% \begin{table}[t]
% \small
% \vspace{-1.em}
% \centering
% % ------------------------------------------------
% \tablestyle{1pt}{1.0}
% \begin{tabular}{x{68}|x{54}|x{54}x{54}c}
% pre-train &
% AP$_\text{50}$ &
% AP &
% AP$_\text{75}$ & \\ 
% \shline
% \randinit & \resrand{10.75}{} & \resrand{}{} & \resrand{}{} & \\
% \supimgnet  & \ressup{14.24}{} & \ressup{}{} & \ressup{}{} & \\
% \hline
% \mocobs & \res{12.45}{-}{0.3} & \reshl{}{+}{0.6} & \reshl{}{+}{0.8} & \\
% \mocobsrs & \res{14.28}{+}{0.2} & \reshl{}{+}{1.5} & \reshl{}{+}{2.1} & \\
% \mocobsrss & \res{17.65}{+}{0.2} & \reshl{}{+}{1.5} & \reshl{}{+}{2.1} & \\
% \mocobsrsst & \res{16.56}{+}{0.2} & \reshl{}{+}{1.5} & \reshl{}{+}{2.1} & \\
% \end{tabular}
% % ------------------------------------------------
% \\
% % ------------------------------------------------
% % end of subfloat
% % ------------------------------------------------
% \vspace{1.0em}
% \caption{Object detection fine-tuned on the xView training set and validated on the test set.
% }
% \label{tab:xview}
% \vspace{-.3em}
% \end{table}

\begin{table}[t]
\small
\vspace{-1.em}
\centering
% ------------------------------------------------
\tablestyle{1pt}{1.0}
\resizebox{0.65\columnwidth}{!}{%
\begin{tabular}{x{106}|x{54}c}
pre-train &
AP$_\text{50}$ $\uparrow$\\ 
\shline
\randinit & \resrand{10.75}{} &\\
\supimgnetpret  & \ressup{14.44}{} &\\
\supimgnet  & \ressup{14.42}{} &\\
\hline
\mocobs & \res{15.45}{+}{4.70}  &\\
\mocobsrs & \res{15.63}{+}{4.88} &\\
\mocobsrss & \res{17.65}{+}{6.90} &\\
\mocobsrsst & \cellcolor{blue!25}\textbf{\res{17.74}{+}{6.99}} &\\
\end{tabular}}
\\
% \vspace{1.0em}
\caption{Object detection results on the xView dataset.
}
\label{tab:xview}
\vspace{-.3em}
\end{table}

% \vspace{-1.0em}
\subsubsection{Image Segmentation}
In this section, we perform downstream experiments on the task of Semantic Segmentation on SpaceNet dataset~\cite{van2018spacenet}. The SpaceNet datasets consists of 5000 high resolution satellite images with segmentation masks for buildings.
% \vspace{-0.6em}
\paragraph{Implementation Details}
We divide our SpaceNet dataset into training and test sets of 4000 and 1000 images respectively. We use PSAnet~\cite{zhao2018psanet} network with ResNet-50 backbone to perform semantic segmentation. We train PSAnet network with a batch size of 16 and a learning rate of 0.01 for 100 epochs and use SGD optimizer.
% \vspace{-0.1em}
\paragraph{Qualitative Analysis}
Table~\ref{tab:spacenet} shows the segmentation performance of differently initialized backbone weights on the SpaceNet test set. Similar to object detection, we achieve the best IoU scores with the addition of temporal positives and geo-location classification task. Our final model outperforms the randomly initialized weights and supervised learning by $3.58\%$ and $2.94\%$ IoU scores. We observe that the gap between the best and worst models shrinks going from the image recognition to object detection, and semantic segmentation task. This aligns with the performance of the MoCo-v2 pre-trained on ImageNet and fine-tuned on the Pascal-VOC object detection and semantic segmentation experiments~\cite{he2020momentum,chen2020improved}.
\begin{table}[!h]
\small
\vspace{-1.em}
\centering
% ------------------------------------------------
\tablestyle{1pt}{1.0}
\resizebox{0.6\columnwidth}{!}{%
\begin{tabular}{x{108}|x{54}c}
pre-train & mIOU $\uparrow$ \\ 
\shline
\randinit &  \resrand{74.93}{} & \\
\iminit  &  \resrand{75.23}{} & \\
\supimgnetpret  &  \ressup{75.61}{} & \\
\supimgnet  &  \ressup{75.57}{} & \\
\hline
\mocobs & \res{78.05}{+}{3.12} & \\
\mocobsrs & \res{78.42}{+}{3.49} & \\
\mocobsrss & \res{78.48}{+}{3.55} & \\
\mocobsrsst &  \cellcolor{blue!25}\textbf{\res{78.51}{+}{3.58}} & \\
\end{tabular}}
\\
% \vspace{1.0em}
\caption{Semantic segmentation results on Space-Net.}
\label{tab:spacenet} 
% \vspace{-.3em}
\end{table}

\begin{table}[!h]
\small
\vspace{-1.em}
\centering
% ------------------------------------------------
\tablestyle{1pt}{1.0}
\resizebox{0.6\columnwidth}{!}{%
\begin{tabular}{x{108}|x{68}c}
pre-train & Top-1 Accuracy $\uparrow$ \\ 
\shline
\randinit &  \resrand{51.89}{} & \\
\iminit  &  \resrand{53.46}{} & \\
\supimgnetpret  &  \ressup{54.67}{} & \\
\supimgnet  &  \ressup{54.46}{} & \\
\hline
\mocobs & \res{55.18}{+}{3.29} & \\
\mocobsrs & \cellcolor{blue!25} \textbf{\res{58.23}{+}{6.34}} & \\
\mocobsrss & \res{57.10}{+}{5.21} & \\
\mocobsrsst &  \res{57.63}{+}{5.74} & \\
\end{tabular}}
\\
% \vspace{1.0em}
\caption{Land Cover Classification on NAIP dataset.}
\label{tab:naip}
% \vspace{-.3em}
\end{table}

% \vspace{-1em}
\subsubsection{Land Cover Classification}
Finally, we perform transfer learning experiments on land cover classification across 66 land cover classes using high resolution remote sensing images obtained by the USDA's National Agricultural Imagery Program (NAIP). We use the images from the California's Central Valley for the year of 2016. Our final dataset consists of 100,000 training and 50,000 test images. 
% More experimental details can be found in \textbf{Appendix}. 
Table~\ref{tab:naip} shows that our method outperforms the randomly initialized weights by $6.34\%$ and supervised learning by $3.77\%$.

\subsection{Experiments on GeoImageNet}
After fMoW, we adopt our methods for unsupervised learning on fMoW for improving representation learning on the GeoImageNet. Unfortunately, since ImageNet does not contain images from the same area over time we are not able to integrate the temporal positive pairs into the MoCo-v2 objective. However, in our GeoImageNet experiments we show that we can improve MoCo-v2 by introducing geo-location classification pre-text task.

\paragraph{Qualitative Analysis}
Table~\ref{tab:geoimagenet} shows the top-1 and top-5 classification accuracy scores on the test set of GeoImageNet. Surprisingly, with only geo-location classification task we can achieve $22.26\%$ top-1 accuracy. With MoCo-v2 baseline, we get $38.51$ accuracy, about $3.47\%$ more than supervised learning method. With the addition of geo-location classification, we can further improve the top-1 accuracy by $1.45\%$. These results are interesting in a way that MoCo-v2 (200 epochs) on ImageNet-1k performs $8\%$ worse than supervised learning whereas it outperforms supervised learning on our \emph{highly imbalanced} GeoImageNet with 5150 class categories which is about $5 \times$ more than ImageNet-1k. 
% We include results on object detection and semantic segmentation on PASCAL VOC dataset in \textbf{Appendix} and show that our method leads to improved performance on these tasks.

\begin{table}[!h]
\resizebox{0.98\columnwidth}{!}{%
\begin{tabular}{@{}c|c|c|c@{}}
\toprule
 & \textbf{Backbone} & \textbf{\begin{tabular}[c]{@{}c@{}}Top-1\\ (Accuracy) $\uparrow$\end{tabular}} & \textbf{\begin{tabular}[c]{@{}c@{}}Top-5\\ (Accuracy) $\uparrow$\end{tabular}} \\ \midrule
\supimgnet & ResNet50 & 35.04 & 54.11\\
\hline
\geoloc & ResNet50 & 22.26 & 39.33 \\
\hline
\baseline & ResNet50 & 38.51 & 57.67 \\
\baselinegeo & ResNet50 & \cellcolor{blue!25} \textbf{39.96} &  \cellcolor{blue!25} \textbf{58.71}\\
\end{tabular}}
\caption{Experiments on GeoImageNet. We divide the dataset into 443,435 training and 100,000 test images across 5150 classes. We train MoCo-V2 and MoCo-V2+Geo for 200 epochs whereas \textbf{Sup. and Geoloc. Learning are trained until they converge}. }
\label{tab:geoimagenet}
\end{table}

% % \subsection{Fairness Experiments on GeoImageNet}
% % \burak{We will check if we get any good news on this.}

% \chenlin{
% }

% \begin{verbatim}
% lr-0.03_bs-256_t-0.02_mocodim-128_image_224_448_original_without_initial_resize_to_224_448 (linear classifier epoch 100): 43.81%
% lr-0.03_bs-256_t-0.02_mocodim-128_image_224_448_comparable_448 (linear classifier epoch 100): 44.12%
% lr-0.03_bs-256_t-0.02_mocodim-128_image_224_448_original_with_initial_resize_to_224 (linear classifier epoch 100): 43.51%
% New:
% lr-0.03_bs-256_t-0.02_mocodim-128_image_224_448_comparable_672 (linear classifier epoch 100):
% 43.583 %  
% \end{verbatim}
\section{Conclusion}
In this work, we provide a self-supervised learning framework for remote sensing data, where unlabeled data is often plentiful but labeled data is scarce. By leveraging spatially aligned images over time to construct temporal positive pairs in contrastive learning and geo-location in the design of pre-text tasks, we are able to close the gap between self-supervised and supervised learning on image classification, object detection and semantic segmentation on remote sensing and other geo-tagged image datasets.

\section*{Acknowledgement}
This research is based upon work supported in part by the Office of the Director of National Intelligence (ODNI), Intelligence Advanced Research Projects Activity (IARPA), via 2021-2011000004. The views and conclusions contained herein are those of the authors and should not be interpreted as necessarily representing the official policies, either expressed or implied, of ODNI, IARPA, or the U.S. Government. The U.S. Government is authorized to reproduce and distribute reprints for governmental purposes not-withstanding any copyright annotation therein.

This research was also supported by Stanford Data for Development Initiative, HAI, IARPA SMART, ONR (N00014-19-1-2145), and NSF grants \#1651565 and \#1733686.

%\input{appendix}

{\small
\bibliographystyle{ieee_fullname}
\bibliography{egbib}
}
\end{document}